\documentclass{ifacconf}

\usepackage{graphicx}      
\usepackage{natbib}        

\usepackage{amsmath,amssymb,amsfonts}
\usepackage{algorithm}
\usepackage{algpseudocode}
\usepackage{comment}
\usepackage{url}
\usepackage{svg}
\usepackage{booktabs}
\usepackage{multirow}
\usepackage[nocomma]{optidef}

\begin{document}
\begin{frontmatter}

\title{Optimized Scheduling and Positioning of Mobile Manipulators in Collaborative Applications} 

\author[First]{Christian Cella} 
\author[First]{Sole Ester Sonnino} 
\author[First]{Marco Faroni}
\author[First]{Andrea Maria Zanchettin}
\author[First]{Paolo Rocco}

\address[First]{Politecnico di Milano, Dipartimento di Elettronica, Informazione e Bioingegneria, Piazza Leonardo da Vinci 32, 20133, Milano (Italy) (e-mail: \{christian.cella, marco.faroni, andreamaria.zanchettin, paolo.rocco\}@polimi.it, soleester.sonnino@mail.polimi.it).}

\begin{abstract}
The growing integration of mobile robots in shared workspaces requires efficient path planning and coordination between the agents, accounting for safety and productivity. In this work, we propose a digital model-based optimization framework for mobile manipulators in human-robot collaborative environments, in order to determine the sequence of robot base poses and the task scheduling for the robot. The complete problem is treated as black-box, and Particle Swarm Optimization (PSO) is employed to balance conflicting Key-Performance Indicators (KPIs). We demonstrate improvements in cycle time, task sequencing, and adaptation to human presence in a collaborative box-packing scenario.
\end{abstract}

\end{frontmatter}

\section{Introduction}
\label{sec:intro}
Human-robot collaboration (HRC) is transforming industrial automation, promoting safer and more efficient workplaces where humans and robots dynamically share tasks. In this context, mobile manipulators are increasingly deployed to address tasks requiring both reachability and flexibility. However, the efficiency of such systems critically depends on optimal path planning and task sequencing, particularly when human interactions are involved. To address these challenges, we propose a general framework for optimizing the sequence of robot base poses and the task scheduling during the \textit{pre-deployment} phase. Our approach optimizes the HRC process before the physical unit is actually deployed, considering robot-specific process parameters and deterministic human trajectories provided by an external \textit{Enterprise Resource Planning} (ERP) system. Because industrial HRC processes can hardly be expressed as analytical cost functions, we formulate the optimization as a \textit{black-box} problem, and we rely on a \textit{digital model} to simulate the overall process and obtain the desired KPIs. To the best of our knowledge, implementing a non-heuristic pipeline to be applied in the \textit{pre-deployment} phase for mobile manipulators in collaborative applications has not yet been addressed in the literature.\\
\begin{figure}
\begin{center}
\includegraphics[width=0.42\textwidth]{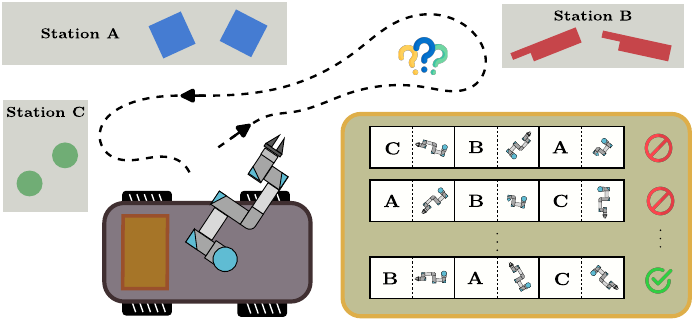}    
\caption{An example of base positioning and scheduling for a mobile manipulator. The robot has to choose the optimal sequence of stations to visit and the optimal base positions for each station.} 
\label{fig:teaser_image}
\end{center}
\end{figure}
As stated by \cite{khatib1999mobile}, a mobile manipulator is defined as a robotic system in which a manipulator is endowed with additional degrees of freedom through a mobile base, seamlessly integrating locomotion with the possibility to interact with the environment. These robotic systems have garnered relevance across various sectors, such as logistics and manufacturing, due to their ability to extend the operational workspace of traditional robotic arms while maintaining advanced manipulation capabilities. In industrial applications, the path planning problem for these units is one of the main challenges that needs to be addressed, especially in the presence of human operators, since it is connected to the optimal placement of the robot base to perform different tasks. In the literature, several approaches have been proposed, focusing on different performance criteria such as maximizing reachability (\cite{yang2009placement}), minimizing cycle time (\cite{wachter2024robot}), or reducing the energy consumption (\cite{gadaleta2017energy}).\\
Working at a stage where the physical unit has not been developed yet entails the adoption of high-fidelity simulators through which it is possible to substitute the evaluation of function with the result of simulations, as proposed by \cite{cella2025digital}. Given the great amount of variables to be considered, the choice of the optimization scheme is of paramount importance. Among the various algorithms, \textit{Particle Swarm Optimization} (PSO) has proven particularly effective for a wide range of problems. Introduced by \cite{kennedy1995particle}, PSO is an evolutionary algorithm which operates by iteratively adjusting the positions of a set of \textit{particles} within a search space, combining individual and collective experiences to converge toward an optimal solution. Its effectiveness has been widely demonstrated in robotic applications, including path planning (\cite{wu2024trajectory}, \cite{masehian2010multi}) and robot base placement (\cite{doan2017optimal}).\\
Although simulation-based methods have been widely applied in robotics, their use for optimizing HRC applications with mobile robots during the \textit{pre-deployment} phase remains an underexplored area. This work aims at filling this gap by proposing a digital model-driven optimization framework capable of jointly determining the optimal sequence of robot base positions and task scheduling, as visible in Fig. \ref{fig:teaser_image}, balancing multiple conflicting objectives through iterative simulations on a \textit{black-box} representation of the process. To verify the validity of the method, we study a \textit{box-packing} scenario where the robot is mounted on a linear guide, while a human operator follows a schedule given by an ERP. In this scenario, we proved that our method is capable of accounting for safety and allows to retrieve the optimal packing conditions.

\section{Problem statement}
\label{sec:problem_statement}
We denote by $\mathcal{T}_r=\{T_{r1},\cdots,T_{rN_R}\}$ the set of all the pick-and-place tasks \textit{P$\&$P} of the robot. We selected \textit{P$\&$P} since it is the most general task for mobile manipulators. At the beginning, $\mathcal{T}_r$ is not known and it must be obtained iteratively as explained in Section \ref{sec:methodology}; however, we know that it must contain $N_R$ tasks $T_{rw}$, $w \in \{1,\cdots,N_R\}$, obtained as the concatenation of the \textit{robotic primitives} present in the set $\mathcal{P}_r$, defined according to the application. Each task $\mathcal{T}_{rw}$ requires a pick point, a place point and the velocity and acceleration at the end-effector ($\mathbf{v_R},\mathbf{a_R}$, possibly scaled as explained in Subsection \ref{sub:presence_operator}). 
Moreover, a motion planner, $\lambda$, computes the movement for the $N_D$ degrees of freedom of the base between each \textit{P$\&$P} according to the maximum velocities and accelerations of the base, $\mathbf{v_B}$ and $\mathbf{a_B}$.\\
To guarantee the generality of the method, we also account for the presence of human operators. Given the optimization stage for which we engineered our method, we supposed to dispose of a \textit{deterministic} plan for the operator's trajectory provided by an \textit{Enterprise Resource Planning} (ERP) system. Starting from a set of available \textit{elementary} human tasks specified in $\mathcal{P}_{h}$, we define the ERP as a function that outputs the set $\mathcal{T}_h=\{T_{h1},\cdots,T_{h N_h}\}$ that contains all the $N_h$ tasks of the operator, $\boldsymbol{\tau_h}=[\tau_{h1},\cdots,\tau_{hN_h}]$ containing the starting times for the tasks and a list $\mathbf{p_h}=[[x_{h1}, y_{h1}, z_{h1}],\cdots,[x_{hN_h},y_{hN_h},z_{hN_h}]]$ that allows to know the exact position of the operator in time. It must be noted that, based on this assumption, $\mathcal{T}_h$, $\boldsymbol{\tau_h}$ and $\mathbf{p_h}$ are known before starting the optimization procedure, unlike $\mathcal{T}_r$ which is updated step by step by our method.\\
The aim of this work is to determine the optimal sequence of base poses $\mathbf{x^*} \in \mathbb{R}^{N_DN_R}$ and the optimal sequence of operations $\boldsymbol{\theta^*} \in \mathbb{N}^{N_R}$ that allow to optimize the HRC process described by a general multi-objective function $\mathbf{f}_{\text{HRC}}:\mathbb{R}^{N_D N_R}\times\mathbb{N}^{N_R} \rightarrow \mathbb{R}^{N_K}$, which is not known a-priori, or may be difficult to evaluate. To cope with this problem, we substitute the analytical resolution of the presented problem with the iterative optimization scheme presented in the following, in which the evaluations of $\mathbf{f}_{\text{HRC}}$ are obtained from a simulated digital twin.

\section{Methodology}
\label{sec:methodology}
Although the proposed method is general, we illustrate its implementation through a box-packing scenario as a representative case study. This choice, without loss of generality, allowed us to introduce a systematic notation to describe the elements involved, facilitating the explanation of the methods we implemented.

\subsection{Notation}
\label{sub:nomenclature_items}

\begin{figure}
\begin{center}
\includegraphics[width=0.38\textwidth]{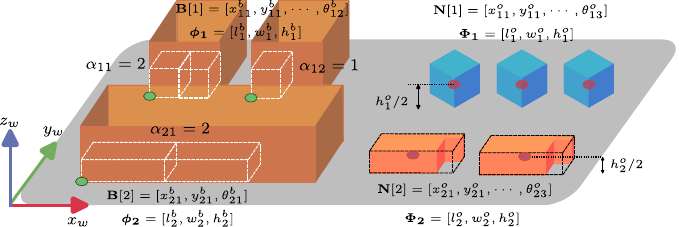}    
\caption{General packing scenario. The objects in blue and orange are the \textit{pick-side} items enclosed in their \textit{convex hulls} (dashed-black lines). Inside each box, the dashed-white elements are the (\textit{place-side} obejcts).} 
\label{fig:general_packing}
\end{center}
\end{figure}

The definitions presented in this subsection are based on the assumption that objects (denoted by superscript $o$) of the $i$-th type possess the same dimensions $\boldsymbol{\Phi_i}=[l_i^o,w_i^o,h_i^o]$, referred to the smallest \textit{convex hull} that encloses them. Moreover, also all the boxes (superscript $b$) available for a specific type of item share the same dimensions $\boldsymbol{\phi_i}=[l_i^b,w_i^b,h_i^b]$. In the remainder of this work, the subscripts $i,j,k,s$ will be the same as the ones used in Algorithms \ref{alg:best_position_optimal_sequence} and \ref{alg:run_pso}.\\ 
We denote by $\mathcal{M}=\{1,2,\cdots,m\}$ the set of the $m$ indices that allow to address a specific item type. Instead, $\mathcal{N}=\{\eta_1,\cdots,\eta_m\}$ is the set whose elements represent the number of items for each type we dispose of. $\mathbf{N}=[[\mathbf{N_{11},\cdots,\mathbf{N_{1 \boldsymbol{\eta_1}}}}],\cdots,[\mathbf{N_{m1},\cdots,\mathbf{N_{m \boldsymbol{\eta_m}}}}]$ is the set of \textit{pick-side} items of each type that have to be packed, with $\mathbf{N}[i][s] \in \mathbb{R}^3$ representing the number of degrees of freedom $\{x, y, \theta\}$ defining the centroid of each item $s$ of type $i$ (red circles in Fig. \ref{fig:general_packing}).\\
For what concerns boxes, we considered them to be fixed in space. $\mathcal{B} = \{\beta_1, \cdots, \beta_m\}$ is the set specifying the number of bins available to pack a generic type of item, while $\mathbf{B}=[[\mathbf{B_{11},\cdots,\mathbf{B_{1 \boldsymbol{\beta_1}}}}],\cdots,[\mathbf{B_{m1},\cdots,\mathbf{B_{m \boldsymbol{\beta_m}}}}]$ contains the information about position and orientation of the $j$-th box of type $i$, with $\mathbf{B}[i][j] \in \mathbb{R}^3$ (the coordinates are referred to the green circles in Fig. \ref{fig:general_packing}).\\
We define as $\alpha_{ij}$ the maximum number of objects of type $i \in \mathcal{M}$ that can be placed inside one of the $j$ available boxes for that type ($j \in \{1,\cdots,\beta_i\}$). Moreover, we define as \textit{place-side} spots the \textit{virtual} poses of the items in the boxes, and we denote them by $\boldsymbol{\gamma_{ij}}=[\boldsymbol{\gamma_{ij,1}}, \cdots, \boldsymbol{\gamma_{ij,\alpha_{ij}}}]$, in which the $k$-th element $\boldsymbol{\gamma_{ij,k}}$ contains the three coordinates ($\{x,y,\theta\}$) of one of the available spots \textit{place-side}.

\subsection{Layout of the place-side objects}
\label{sub:best_fit_algorithm}
The first problem to be solved concerns the position of the \textit{place-side} objects. This issue has been solved through the implementation of the \textit{Best-fit algorithm} described in \cite{dube2006optimizing}, which allows to retrieve two quantities. The first one is $\boldsymbol{\alpha}=[[\alpha_{11},\cdots,\alpha_{1\beta_1}],\cdots,[\alpha_{m1},\cdots,\alpha_{m\beta_m}]]$, whose elements are lists containing the quantities $\alpha_{ij}$ defined in Section \ref{sec:problem_statement}, while the second output is $\boldsymbol{\Gamma} = [[\boldsymbol{\gamma_{11}},\cdots, \boldsymbol{\gamma_{1 \beta_1}}],\cdots,[\boldsymbol{\gamma_{m1}},\cdots, \boldsymbol{\gamma_{m \beta_m}}]$, whose elements are lists containing the elements $\boldsymbol{\gamma_{ij}}$ (coordinates of the \textit{place-side} objects defined in Subsection \ref{sub:nomenclature_items}). 
For the specific problem addressed in this work we modified the original implementation of the \textit{Best-fit algorithm} proposed in the GitHub repository \texttt{3dbinpacking}\footnote{The repository can be found at: \url{https://github.com/enzoruiz/3dbinpacking.git}}, in order to adapt it to our case. More in detail, each \textit{P$\&$P} $\mathcal{T}_{rw}$ is performed with a suction cup and the problem is \textit{planar}; for this reason, the feasible rotations of the items in the boxes ($\theta$ values of the \textit{place-side} items) can only be around the axis $z_w$ (see Fig. \ref{fig:general_packing}) and we only considered $\theta=0$ or $\theta=\pi/2$ as possible rotations.

\subsection{Optimal base poses and sequencing strategy}
\label{sub:base_and_sequencing}

The second section of the method we implemented allows to retrieve the sequence of optimal poses of the robot base $\mathbf{x}^*$ and the optimal sequence of tasks $\boldsymbol{\theta^*}$. Building on the lists $\boldsymbol{\Gamma}$ and $\boldsymbol{\alpha}$, all the possible types of items ($i \in \mathcal{M}$) are scanned one at a time, as visible in Algorithm \ref{alg:best_position_optimal_sequence} at step \ref{step:scan_item_types}. For all the $j$ boxes ($j \in \{1,\cdots, \beta_i\}$) that must be filled with items of type $i$, we know how many items ($\alpha_{ij}$) can be positioned in that specific box in the \textit{place-side} spots. For this reason, we evaluate for every $k$ \textit{place-side} position all the \textit{pick-side} items that have not been packed yet and, for all of them, we obtain the optimal sequence of the robot base poses (defined as $\mathbf{\bar{x}}=[[x_{11},\cdots,x_{1N_D}],\cdots,[x_{\eta_i1},\cdots,x_{\eta_iN_D}]]$ to account for $N_D$ degrees of freedom) and the associated times to perform the \textit{P$\&$P}, together with the vector of scaling coefficients $\boldsymbol{\kappa}$ used in the process.\\
Subsequently, among all the possible \textit{P$\&$P} tasks, the selected operation is the one associated to the item $k$ that resulted in the optimal base pose $\mathbf{\bar{x}}[\eta_i^*][:]$ that is closer to the current one of the robot $\mathbf{x_0}$ (steps \ref{step:travel_1} and \ref{step:travel_2}, described more in detail in Subsection \ref{subsub:minimze_travel_time}).\\
As previously anticipated, our method takes into account the presence of human operators in the environment: as it can be seen in step \ref{step:append_time} of Algorithm \ref{alg:best_position_optimal_sequence}, it is possible to update the current time $t_c$, as the sum of the times required to perform the \textit{P$\&$P} task ($\mathbf{T_{pp}}^*[\eta^*_i]$) and the time $c^*$ to move the robot base to the newly computed position, according to the planner $\lambda$. As explained in Subsection \ref{subsub:pso_explanation}, the evaluation of $t_c$ is of paramount importance to implement the \textit{look ahead} approach to scale the robot velocity and enforce the safety measures.

\subsubsection{Particle Swarm Optimization (PSO)}
\label{subsub:pso_explanation}
To obtain the best pose of the robot base for each \textit{pick-side} item not yet packed (items present in $\mathcal{C}$ in step \ref{step:set_C_1} of Algorithm \ref{alg:run_pso}) we leveraged PSO. In our work, each one of the $N_p$ particles of the swarm represents a candidate pose for the manipulator's base. The PSO evolution works by iteratively updating the position of the particles $\mathbf{p_{pos}}$ based on their fitness $\boldsymbol{\psi}$, that we obtained as the normalized sum of $N_K=2$ KPIs and two additional corrective factors. More in detail (step \ref{step:KPIs}), for each simulation we stored in $\mathbf{T_{pp}}$ the time to perform the $N_p$ \textit{P$\&$P} tasks, while $\boldsymbol{\delta}$ contains the \textit{inverse manipulability index}: each element is computed as $\bar{\delta}_q = 1/\sqrt{|J J^T|},\ q \in \{1,\cdots,N_p\}$ and represents the average of all the values obtained at time intervals $\Delta t$ during each simulation. In order to sum together values presenting different quantities, $\mathbf{T_{pp}}$ is substituted with $\mathbf{z_t} = (\mathbf{T_{pp}}[p,:] - \mu_{t,i}) /\sigma_{t,i}$
while $\boldsymbol{\delta}$ becomes $\mathbf{z_{\boldsymbol{\delta}}} = (\mathbf{\boldsymbol{\delta}}[p,:] - \mu_{\delta,i})/\sigma_{\delta,i}$ (method called \texttt{Normalization}, applied at the $p$-th iteration of the PSO), where the values $\mu_{t,i}$, $\sigma_{t,i}$, $\mu_{\delta,i}$, $\sigma_{\delta,i}$ are the mean and the standard deviations for the KPIs, calculated (together with $\mu_{\text{TVP}}$) on the datasets explained in Subsection \ref{sub:rand_baseline}. We computed the fitness function as it can be seen in step \ref{step:fitness}, where $\mathbf{w}=[w_t, w_{\delta}]$ is the set of weights, while $\boldsymbol{\xi_C}$ and $\boldsymbol{\xi_F}$ are discrete values that should be 0, but they are set to $+\infty$ as a function of the occurrence of collisions (the former) or an unfaesible layout (the latter).\\
The suggestion of new particle positions $\mathbf{p_{pos}}$ happens iteratively in step \ref{step:pos_update} (with the method \texttt{clip} to enforce the boundaries) and is based on the update of the particles velocities reported in steps \ref{step:inertia} and \ref{step:vel_update}: the \textit{cognitive} component $c_1 r_1 (\mathbf{p_{pb}}-\mathbf{p_{pos}})$ depends on the \textit{personal best} $\mathbf{p_{pb}}$ obtained at step \ref{step:pers_best_update}, while the \textit{social} term $c_2 r_2 (\mathbf{p^*} - \mathbf{p_{pos}})$ is a function of the global best $\mathbf{p^*}$ obtained at step \ref{step:glob_best_update} ($r_1$ and $r_2$ are two random numbers $\in[0,1]$ that guarantee the controlled exploration of the search space). Moreover, based on the definition proposed in \cite{shi1998modified}, we also accounted for the \textit{inertial} term $\omega\cdot\mathbf{p_{pos}}$ that depends on a decreasing inertial hyper-parameter $\omega = \omega_0 + p (\omega_n-\omega_0) / N_s$.

\subsubsection{Optimization of the travel time}
\label{subsub:minimze_travel_time}
As soon as $\mathbf{x}$ and $\mathbf{T_{pp}^*}$ are returned by \texttt{RunPSO}, it is possible to determine which item of type $i$ should be picked next (step \ref{step:travel_2} in Algorithm \ref{alg:best_position_optimal_sequence}). For this purpose, we select the item, whose index is called $\eta_i^*$, that is associated with the shortest travel time $c^*$, computed based on the maximum velocity $\mathbf{v_B}$ and maximum acceleration $\mathbf{a_R}$ of the mobile manipulator. The vector of travel times $\mathbf{C}_{\lambda}$ is computed considering as start pose the current one of the robot base $\mathbf{x_0}$, while the goal pose is, one at a time, each element of $\mathbf{x}$ (we suppose a planner, $\lambda$, implements policies to avoid collisions with the human operators during the motion). Consequently, at step \ref{step:append} the optimal robot base poses $\mathbf{x^*}$ and the optimal sequence of items $\boldsymbol{\theta^*}$ can be obtained recursively. In the end, at step \ref{step:robot_set} the set of robotic \textit{P$\&$P} operations is updated: each element of $\mathcal{T}_{rw}$ will be called \textit{P$\&$P}$_{i\eta_i^*}$, since associated to the item $\eta_i^*$ of type $i$.

\begin{algorithm}
\scriptsize
\caption{Optimal base pose and items sequence}
\label{alg:best_position_optimal_sequence}
\begin{algorithmic}[1]
\State \textbf{Given:} $\mathcal{M}$, $\mathcal{B}$, $\mathcal{N}$, $\textbf{B}$, $\textbf{N}$, $\mathbf{v_R}$, $\mathbf{a_R}$, $\mathbf{v_B}$, $\mathbf{a_B}$, $\lambda$
\State \textbf{Initialize:} $\mathbf{x}^*[:][:] \gets 0$, $\boldsymbol{\theta}^* \gets []$, $\boldsymbol{\kappa}^* \gets []$, $\mathbf{x_0} \gets \mathbf{0}$, $t_c \gets 0$, $w \gets 1$
\For{i $\in \mathcal{M}$} \label{step:scan_item_types}
    \State $t_w \gets \textbf{t}_w[i]$, $\mathcal{A} \gets \emptyset$, r $\gets 0$
    \For{j $\gets 1, \cdots, \beta_i$}
        \State c$^* \gets \infty$, $\alpha_{i,j} \gets \boldsymbol{\alpha}$[i][j]
        \For{k $\gets 1, \cdots, \alpha_{i, j}$}
            \State $\boldsymbol{\gamma} \gets \Gamma$[i][j][k], r $\gets$ r+1
            \State $\mathbf{\bar{x}}, \textbf{T}^*_{\text{pp}}, \boldsymbol{\kappa} \gets \texttt{RunPSO}$(\textbf{N}[i], $\boldsymbol{\gamma}$, $\mathcal{A}$, $t_c$, $t_w$) \label{step:run_pso_method} 
            \State $\textbf{C}_{\lambda} \gets \texttt{ComputeTravelTime}$($\mathbf{\bar{x}}$, $\mathbf{x_0}$, $\mathbf{v_B}$, $\mathbf{a_B}$, $\lambda$) \label{step:travel_1} 
            \State c$^*$, $\eta_i^* \gets \text{min}(\textbf{C}_{\lambda})$ \label{step:travel_2}
            \State $\mathbf{x^*}$[r][:] $\gets \mathbf{\bar{x}}[\eta_i^*][:]$, $\boldsymbol{\theta}^*$[r] $\gets \eta_i^*$, $\boldsymbol{\kappa}^*[r] \gets \boldsymbol{\kappa}[\eta^*_i]$, $\mathcal{A}[r] \gets \eta_i^*$ \label{step:append}
            \State $\mathbf{x_0} \gets \mathbf{x^*}$[r][:], $t_c \gets t_c + \text{c}^* + \textbf{T}^*_{\text{pp}}[\eta_i^*]$ \label{step:append_time}
            \State $\mathcal{T}_r\{w\} \gets \texttt{P\&P}(\mathcal{P}_r, \mathbf{N}[i][\eta_i^*],\boldsymbol{\gamma}, \boldsymbol{\kappa^*}[r], \mathbf{x_0}, \mathbf{v_R}, \mathbf{a_R}, \mathbf{v_B}, \mathbf{a_B})$ \label{step:robot_set}
            \State $w \gets w + 1$
        \EndFor
    \EndFor
\EndFor
\State \textbf{Return:} $\mathbf{x^*}$, $\boldsymbol{\theta^*}$, $\boldsymbol{\kappa}^*$
\end{algorithmic}
\end{algorithm}

\begin{algorithm}
\scriptsize
\caption{\texttt{RunPSO} procedure}
\label{alg:run_pso}
\begin{algorithmic}[1]
\State \textbf{Input:} $\mathbf{N}[i]$, $\boldsymbol{\gamma}$, $\mathcal{A}$, $t_c$, $t_w$
\State \textbf{Given:} $N_p$, $N_s$, $\mathbf{p_{lim}}$, $\mathbf{v_{lim}}$, $\mathbf{w}$, $\omega_o$, $\omega_n$, $c_1$, $c_2$, $\boldsymbol{\tau_h}$, $\mathbf{p_h}$
\State \textbf{Initialize:} $\mathbf{T_{pp}} \gets []$, $\mathbf{T_{pp}^*} \gets []$, $\boldsymbol{\delta} \gets []$, $\boldsymbol{\xi_F} \gets []$, $\boldsymbol{\xi_C} \gets []$, $\mathbf{x}[:][:] \gets \infty$, $\boldsymbol{\kappa} \gets []$, $\mathcal{C} \gets \{1,\cdots,\eta_i\} \setminus \mathcal{A}$ \label{step:set_C_1}
\For{$s \in \mathcal{C}$} \label{step:set_C_2}
    \State $\mathbf{p_{pos}} \gets \texttt{rand}(\mathbf{p_{lim}}, N_p, N_D)$, $\mathbf{p_{vel}} \gets \texttt{rand}(\mathbf{v_{lim}}, N_p, N_D)$
    \State $\kappa \gets \texttt{RecedingHorizon} (t_c, t_w, \boldsymbol{\tau_h}, \mathbf{p_h}, \mathbf{N}[i][s], \boldsymbol{\gamma})$ \label{step:scaling}
    \State $\psi_0 \gets +\infty$, $\mathbf{p^*} \gets \mathbf{0}$, $\mathbf{p_{pb}} \gets \mathbf{p_{pos}}$, $\boldsymbol{\bar{\psi}} \gets [+\infty]$
    \For{$p \gets 1,\cdots,N_s$}
        \State $\mathbf{T_{pp}}[p, :]$, $\boldsymbol{\delta}[p, :]$, $\boldsymbol{\xi_C}[p, :], \boldsymbol{\xi_F}[p, :] \gets \texttt{Sim}(\mathbf{N}[i][s], \boldsymbol{\gamma}, \kappa, \mathbf{p_{pos}})$ \label{step:KPIs}
        \State $\mathbf{z_t}$, $\mathbf{z_{\boldsymbol{\delta}}} \gets \texttt{Normalization}(\mathbf{T_{pp}}[p,:], \boldsymbol{\delta}[p, :])$ \label{step:normalization}
        \State $\boldsymbol{\psi} \gets w_t \mathbf{z_t}+w_{\delta} \mathbf{z_{\boldsymbol{\delta}}} + \boldsymbol{\xi_F}[p, :] + \boldsymbol{\xi_C}[p, :]$ \label{step:fitness}
        \State $\mathbf{p_{pb}} \gets \mathbf{p_{pos}}[\boldsymbol{\psi} < \boldsymbol{\bar{\psi}}][:]$ \label{step:pers_best_update}
        \State $\boldsymbol{\bar{\psi}} \gets \boldsymbol{\psi}[\boldsymbol{\psi}<\boldsymbol{\bar{\psi}}]$
        \If{$\text{min}(\boldsymbol{\psi})<\psi_0$}
            \State $h \gets \text{argmin}(\boldsymbol{\psi})$
            \State $\mathbf{p^*} \gets \mathbf{p_{pos}}[h][:], \psi_0 \gets \boldsymbol{\psi}[h], \bar{T}_{pp} \gets \mathbf{T_{pp}}[p, h]$ \label{step:glob_best_update}
        \EndIf
        \State $r_1, r_2 \gets \texttt{rand}(0, 1, N_p)$, $\omega \gets \omega_0 + p(\omega_n-\omega_0)/N_s$ \label{step:inertia}
        \State $\mathbf{p_{vel}} \gets \omega\cdot\mathbf{p_{vel}} + c_1 r_1(\mathbf{p_{pb}}-\mathbf{p_{pos}}) + c_2 r_2 (\mathbf{p^*}-\mathbf{p_{pos}}) $ \label{step:vel_update}

        \State $\mathbf{p_{pos}} \gets \mathbf{p_{pos}} + \mathbf{p_{vel}}$, $\mathbf{p_{pos}} \gets \texttt{clip}(\mathbf{p_{pos}}, \mathbf{p_{lim}})$ \label{step:pos_update}
    \EndFor
    \State $\mathbf{\bar{x}}[s][:] \gets \mathbf{p^*}, \mathbf{T_{pp}^*}[s] \gets \bar{T}_{pp}$, $\boldsymbol{\kappa}[s]\gets \kappa$
\EndFor
\State \textbf{Return:} $\mathbf{x}$, $\mathbf{T_{pp}^*}$, $\boldsymbol{\kappa}$
\end{algorithmic}
\end{algorithm}

\begin{figure*}
\begin{center}
\includegraphics[width=0.95\textwidth]{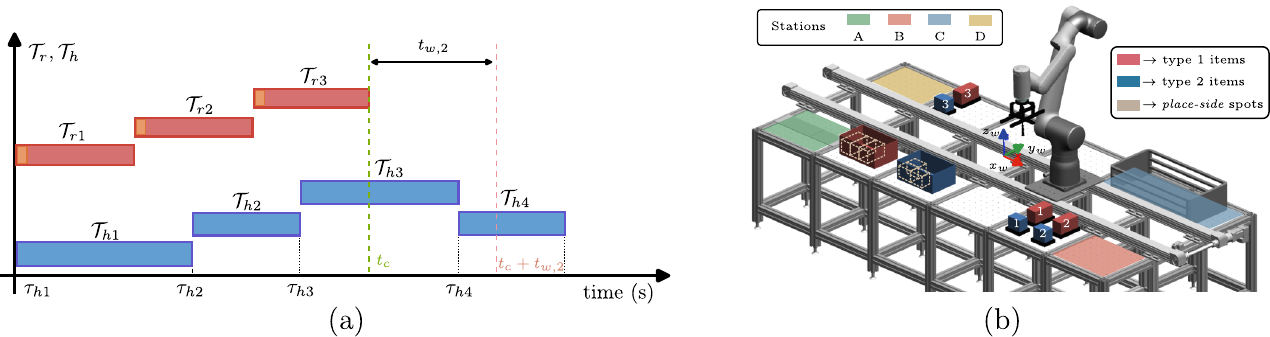}
\caption{a) Example of a Gantt chart representing both the robot tasks $\mathcal{T}_r$ (goal of the optimization) and the human schedule $\mathcal{T}_h$ (given by an external ERP system). b) Layout considered in the use case.} 
\label{fig:receding_horizon}
\end{center}
\end{figure*}

\subsection{Presence of the human operator}
\label{sub:presence_operator}
Whenever a HRC unit is setup, the operator's safety must be guaranteed according to the ISO-TS/15066 norm (\cite{ISOTS15066}). In this work, we implemented safety by modulating the speed and the acceleration at the robot end-effector based on the minimum distance of the human with respect to current \textit{pick-side} item or \textit{place-side} spot, whichever is closer (this safety logic works during \textit{P$\&$P} tasks; the one of \textit{collision-avoidance} implemented in the planner $\lambda$ during the motion of the mobile base). For this purpose, we define $n_k$ scaling factors contained in a set $\mathcal{K}=\{0,\cdots,n_k-1\}$. The selected scaling value, called $\kappa$, specifies how severe the modulation of the maximum values of the TCP parameters is for the \textit{place-side} spot $k$ and for the \textit{pick-side} item $s$, according to

\begin{equation} 
\label{eq:scaling}
\{\mathbf{v},\mathbf{a}\}_{ks}=\{\mathbf{v},\mathbf{a}\}_{\mathbf{R}}- \kappa\frac{1}{(n_k-1)} \{\mathbf{v},\mathbf{a}\}_{\mathbf{R}}
\end{equation}

As visible in Fig. \ref{fig:receding_horizon}a, knowing the cumulative packing time $t_c$ (updated at step \ref{step:append_time} in Algorithm \ref{alg:best_position_optimal_sequence}) is not sufficient, since the scheduling given by the ERP system for the human may require them to move to new locations (i.e. to perform the task $\mathcal{T}_{h4}$), while the robot performs $\mathcal{T}_r\{4\}$. The problem is that the time for $\mathcal{T}_r\{4\}$ is not known yet, since it is the goal of the current optimization step, and this may hinder the applied safety measure: if in $\mathcal{T}_{h3}$ the operator is far from the line, but in $\mathcal{T}_{h4}$ the human is close to the robot, considering only the task $\mathcal{T}_{h3}$ leads to unsafe conditions.\\
To solve this problem, we introduced a method that ``looks ahead'' of $t_{w,i}$ seconds. A reasonable time window to implement this approach is $t_{w,i} = \mu_{t,i} + \mu_{\text{TVP}}$, where $i$ is the index related to the specific type of item under consideration, while $\mu_{t,i}$ and $\mu_{\text{TVP}}$ are the parameters introduced in Subsection \ref{subsub:pso_explanation}. The rationale behind $t_{w,i}$ is the following: since we do not know how much time is going to take to perform the new \textit{P$\&$P} task under exam, we approximate its duration with the time that the \textit{P$\&$P} is probably going to last. At this point, we generate a set $\mathcal{Q} = \{q_1,\cdots,q_n\}$ containing the indices of the $n$ human tasks happening in the time window $t_{w,i}$ (for instance, in Fig. \ref{fig:receding_horizon}a, $\mathcal{Q}=\{3,4\}$) and for all $c$ elements of $\mathcal{Q}$ we compute two distances: $d_{\text{I},c}= |\mathbf{p_h}[c]-\mathbf{N}[i][j]|$ for \textit{pick-side} items; $d_{\text{II},c}= |\mathbf{p_h}[c] - \boldsymbol{\Gamma}[i][j][k]|$ for \textit{place-side} spots. Once all the distances have been computed, the minimum one, called $d_{min}$ is going to be compared with the safety distances encoded in $\mathbf{d_{s,min}}$ and $\mathbf{d_{s,max}}$ (both $\in \mathbb{R}^{n_k}$). The index $v\in\{1,\cdots,n_k\}$ of the couple of values $\mathbf{d_{s,min}}[v] - \mathbf{d_{s,max}}[v]$ containing $d_{min}$ allows to retrieve the correct value of $\kappa=\mathcal{K}\{v\}$ to scale $\mathbf{v_R}$ and $\mathbf{a_R}$ according to equation \eqref{eq:scaling}.

\section{Experimental validation}
\label{sec:experiment}
\begin{figure*}
\begin{center}
\includegraphics[width=0.99\textwidth]{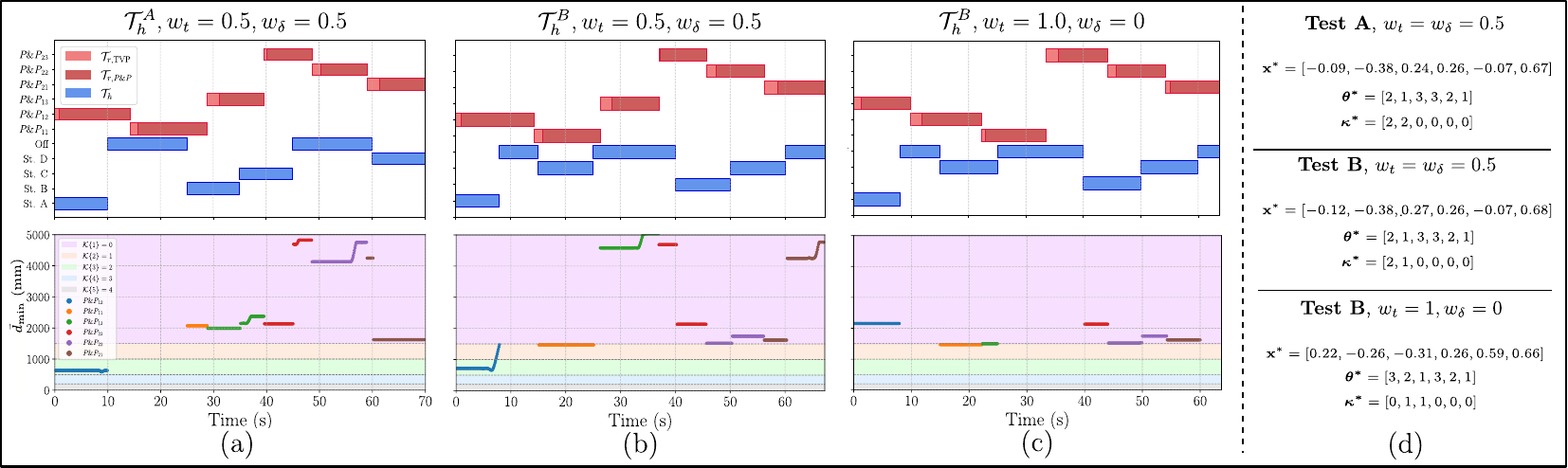}
\caption{Results of the tests performed. a) Test A with equal importance of the KPIs. b) Test B with equal importance of the KPIs. c) Test B prioritizing time. d) $\mathbf{x^*}, \boldsymbol{\theta^*}, \boldsymbol{\kappa^*}$ for the tests.} 
\label{fig:results}
\end{center}
\end{figure*}

For the experimental validation we used the ABB CRB 15000 cobot, characterized by 6 degrees of freedom, a maximum payload of 12 Kg, $\mathbf{v_R}=2$ m/s and $\mathbf{a_R}=27$ m/s$^2$. We mounted the robot on a linear axis ($N_D=1$), which allowed us to search for the optimal base positions on a length of 3 m ($\pm 1.5$ m with respect to the reference frame in Fig. \ref{fig:receding_horizon}b). The planning system $\lambda$ generates linear trajectories between the start pose and the goal, and it implements a \textit{Trapezoidal Velocity Profile} (TVP), leveraging the following parameters: $\mathbf{v_B}=0.7$ m/s and $\mathbf{a_B}=0.5$ m/s$^2$. All the simulations were executed by leveraging a socket communication between python and the C$\#$ APIs of Tecnomatix Process Simulate 2307 by Siemens, after selecting the parameters visible in Table \ref{table:PSO_parameters}. In terms of hardware, we used a Legion Pro 5 16IRX9 laptop, featured with a Intel(R) Core(TM) i9-14900HX  processor operating at 2.20 GHz and 32 GB of RAM.\\
The planar packing scenario is represented in Fig. \ref{fig:receding_horizon}b: we chose $m=2$ types of items, each composed by the same number of items ($\eta_1=\eta_2=3$), characterized by $\boldsymbol{\Phi_1}=[0.155, 0.085, 0.085]$ m and $\boldsymbol{\Phi_2} = [0.105, 0.08, 0.085]$ m. We also supposed to have only one box available for each object type ($\beta_1=\beta_2=1$) and $\boldsymbol{\phi_1} = \boldsymbol{\phi_2} = [0.285,0.19, 0,13]$ m. For the scaling of the velocity enforced through equation \eqref{eq:scaling}, we selected $n_k=5$ possible reductions ($\mathcal{K}=\{0,1,2,3,4\}$), as a function of the safety region whose limits are $\mathbf{d_{s,max}} = [+\infty, 1.5, 1, 0.5, 0.2]$ m and $\mathbf{d_{s,min}} = [1.5, 1, 0.5, 0.2, 0]$ m.\\
Each \textit{P$\&$P} operation $\mathcal{T}_{rw}$ ($w \in \{1,\cdots,N_R=6\}$) is created by concatenating the primitives contained in $\mathcal{P}_r$ = \{\textit{MoveTo, Overfly, SuctionOn, SuctionOff, MoveBase}\}, as it can be seen in the accompanying video\footnote{The video is available at: \url{https://youtu.be/Q5XaGDUOH0M}}. Instead, for the human $\mathcal{P}_h =$ \{\textit{CreateBox}$_{A}$, \textit{BringItems}$_{B}$, \textit{FillPallet}$_{C}$, \textit{BringItems}$_{D}$, \textit{Offline}\} (the subscripts $A,B,C,D$ are referred to the station on the line where the human performs the task, and are abbreviated as ``St.$i$'' in Fig. \ref{fig:results}).
\begin{table}[h]
    \centering
    \caption{Main parameters used in the PSO.}
    \small 
    \resizebox{0.48\textwidth}{!}{ 
    \begin{tabular}{cccccccccc}
        \toprule
        $N_s$ & $N_p$ & $\mathbf{p_{lim}} (m)$ & $\mathbf{v_{lim}} (m/s)$ & $w_t$ & $w_{\delta}$ & $\omega_0$ & $\omega_n$ & $c_1$ & $c_2$ \\
        \toprule
        25 & 10 & [-1.5, 1.5] & [-1, 1] & 0.5 & 0.5 & 0.9 & 0.4 & 2 & 2 \\
        \bottomrule
    \end{tabular}
    } 
    \label{table:PSO_parameters}
\end{table}

\subsection{Creation of the inference dataset}
\label{sub:rand_baseline}

The fitness function $\boldsymbol{\psi}$ accounts for the time to execute a specific \textit{P$\&$P} and the mean manipulability. As explained in Subsection \ref{subsub:pso_explanation}, these quantities have to be normalized exploiting the mean value and the standard deviation of a certain distribution. For this reason, we executed $N_T=2000$ tests for both the object types. At every iteration, we randomly selected one of the \textit{pick-side} items and one of the \textit{place-side} spots in the only box available for that type, and, in case of feasibility ($\boldsymbol{\xi_F}=0$) and no collisions ($\boldsymbol{\xi_C}=0$), we increased the datasets of time and manipulability. For both of the types, $\approx 800/2000$ tests turned out to be successful. All the tests were performed with the maximum values of $\mathbf{v_R}$ and $\mathbf{a_R}$, without any scaling. The insight behind this choice concerns the need of scaling down $\mathbf{v_R}$ and $\mathbf{a_R}$ as a function of the human proximity to the working zone. In the cases where the parameters reduction is needed, the time is statistically higher than the mean value; as a consequence, the associated value of $\mathbf{z_t}$ computed with equation \eqref{eq:scaling} is higher than the $\mathbf{z_t}$ value for the operations that did not require the parameters scaling. For this reason, the algorithm, during the minimization of $\boldsymbol{\psi}$, tends to converge to solutions that try to avoid reducing $\mathbf{v_R}$ and $\mathbf{a_R}$ as much as possible, increasing the productivity.\\
We adopted a similar approach to also compute the mean value for the motion of the base according to a \textit{Trapezoidal Velocity Profile} (TVP), executing $N_t$ simulations with random initial and goal position. The results are reported in Table \ref{table:baseline_results}.

\begin{table}[h]
    \centering
    \caption{$\mu$ and $\sigma$ obtained from the baseline.}
    \small 
    \resizebox{0.48\textwidth}{!}{ 
    \begin{tabular}{cccccccccc}
        \toprule
        $N_T$ & $\mu_{t,1}$ & $\mu_{t,2}$ & $\mu_{\delta,1}$ & $\mu_{\delta,2}$ & $\sigma_{t,1}$ & $\sigma_{t,2}$ & $\sigma_{\delta,1}$ & $\sigma_{\delta,2}$ & $\mu_{\text{TVP}}$ \\
        \toprule
        2000 & 9.3 $s$ & 8.9 $s$ & 7.4 $s$ & 11.4 $s$ & 0.5 & 0.5 & 1.5 & 3.5 & 2.1 $s$ \\
        \bottomrule
    \end{tabular}
    } 
    \label{table:baseline_results}
\end{table}

\subsection{Results}
\label{sub:tests}
To validate our algorithm, we tested two different sets of tasks for the human operator, called $\mathcal{T}_{h}^A$ and $\mathcal{T}_{h}^B$ and visible as the blue segments in the upper Gantt charts in Fig. \ref{fig:results}. For Test A, according to equation \eqref{eq:scaling}, the first \textit{P$\&$P} should be executed at $50\%$ of the maximum speed and acceleration of the robot ($\boldsymbol{\kappa}[1]=2$). This is also visible from the lower plot in Fig. \ref{fig:results}a that represents the trend of the minimum distance $d_{\text{min}}$: in the first 10 seconds, $d_{\text{min}}$ is $\approx 0.63$ m, and is located in the light-green region that is characteristic of $\mathbf{d_{s,min}}[3]=0.5$ m and $\mathbf{d_{s,max}}[3]=1$ m (resulting in $\mathcal{K}\{3\}=2$). Instead, $\boldsymbol{\kappa}[2]=2$ is less intuitive and requires an additional analysis. After \textit{P$\&$P}$_{12}$ has been determined, $t_c \gets 14.24 s$ and the Algorithm \ref{alg:run_pso} at step \ref{step:scaling} will select a value of $\kappa$ by considering all the possible human tasks in between $14.24s$ and $25.64s$ ($t_{w,1}=\mu_{t,1}+\mu_{\text{TVP}}$, whose values are visible in Table \ref{table:baseline_results}). In this time span, $\mathcal{Q}=\{2,3\}$ and the minimum distance $d_{min}=d_{I,3}$, between the human in station B and the \textit{pick-side} item number 1 of type 1. Despite the human stays in station B for a small amount of time during the execution of the \textit{P$\&$P} to be considered as $\mathcal{T}_r\{2\}$ ($\approx 0.64s$, since the task $\mathcal{T}_{h}\{3\}$ starts at $25s$), this is sufficient to scale down of $50\%$ the velocity and the acceleration for that operation. All the other values contained in $\boldsymbol{\kappa}$ can be deduced by the plots in Fig. \ref{fig:results}a.\\
Fig. \ref{fig:results}b shows the results of Test B, which allowed us to demonstrate both the generality of the procedure (we supposed a different sequence of tasks in $\mathcal{T}_{h,2}$) and also the importance of selecting the weights $w_t$ and $w_{\delta}$ of the KPIs. As it can be seen in Fig. \ref{fig:results}c, the fitness $\boldsymbol{\psi}$ just accounts for the time and the algorithm succeeds in finding a plan that allows to execute the packing action in $63.7s$, against the $68.1s$ of the case with both the weights equal to 0.5. The main difference is given by the fact that the PSO algorithm, prioritizing the time, learns that packing the item number 3 as the first one allows to not reduce the velocity for the first \textit{P$\&$P} operation (this is also confirmed by the associated $\bar{d}_{\text{min}}$, located in the light-pink region characteristic of $\mathcal{K}\{1\}=0$). By comparing the sets $\boldsymbol{\kappa}$ for tests B (see Fig. \ref{fig:results}d), it can be easily understood that the method we devised is capable of finding the optimal solution also in cases where we want to prioritize a single KPI.\\
Additionally, we validated the effectiveness of the described \textit{learning} behavior by implementing a randomized baseline.
The baseline method is obtained by randomizing the particles' position updates (i.e., $\mathbf{p_{pos}} \gets \texttt{rand}(\mathbf{p_{lim}}, N_p)$) so that the algorithm finds feasible, yet suboptimal, solutions. 
For the comparison between this \textit{blind search} method and Test A, we focused on the minimum values of $\psi_0$ achieved  by the methods in any of the $\sum_{i=1}^m \sum_{s=1}^{\eta_i} s$ executions of the \texttt{RunPSO} routine, each running for $N_s=25$ simulations. 
The best result achieved by the random baseline is $\psi^*_{0,\text{rand}}=1.12$, 
which is way worse than the best solution achieved by our method ($\psi^*_{0,A}=-2.02$). 
As a consequence, the total time of the packing process for Test A is $\approx 70s$ for the proposed PSO and $\approx 78s$ for the \textit{random search} (Fig. \ref{fig:results}a). 
To obtain approximately the same results of Test A, the random method should be executed for $N_s=100$, resulting in $\approx 1000$ minutes of computational time (versus the $\approx 200$ of the PSO).\\
Finally, the method we implemented allows to study cases where only the robot is present (\textit{human out of the loop} scenarios). This can easily be obtained by imposing $\mathcal{T}_h = \{Offline\}$ ($\mathcal{Q}=\{1\}$ always), $\mathbf{p_h}=[[+\infty, +\infty]]$ and $\boldsymbol{\tau_h}=[0]$. This simple change in the process parameters allows to focus on conditions that do not require a reduction in the robot speed and acceleration according to equation \eqref{eq:scaling}. The test we performed confirms the validity of our method, both in terms of results ($\boldsymbol{\theta}^*=[0,1,2,2,1,0]$, $\boldsymbol{\kappa^*}=[0,0,0,0,0,0]$) and in terms of time taken to perform the insertion of the items in the boxes ($\approx 58$ seconds, with a reduction of $\approx 8\%$ with respect to the best case with the human).

\section{Conclusion}
\label{sec:conclusion}
This work presents a simulation-based framework for the optimization of base poses and task sequencing for mobile manipulators operating in collaborative scenarios. The proposed method allows to integrate deterministic human motion trajectories into the planning phase, ensuring safety. The underlying optimization problem was formulated as a black-box search, and was solved leveraging the learning-based nature of PSO. Experimental validation demonstrated the effectiveness of the method in generating efficient task sequences, minimizing cycle times, and dynamically adapting robot behavior in response to human proximity constraints. Future research will focus on extending the method to more complex HRC scenarios, integrating uncertainty in the human trajectories.

\begin{ack}
This study was carried out within the MICS (Made in Italy – Circular and Sustainable) Extended Partnership and received funding from Next-Generation EU (Italian PNRR – M4 C2, Invest 1.3 – D.D. 1551.11-10-2022, PE00000004). CUP MICS D43C22003120001.
The authors thank Camozzi S.p.A. for their support to the design of the experimental case study.
\end{ack}

\bibliography{ifacconf}

\end{document}